\documentclass{article}

\usepackage{microtype}
\usepackage{subfigure}
\usepackage{booktabs} 

\usepackage{hyperref}
\usepackage{enumerate}
\usepackage{url}
\usepackage{amsmath}
\usepackage[draft]{fixme}
\usepackage{graphicx}
\usepackage{xcolor}
\usepackage{tikz}
\usepackage{multicol}
\usepackage[utf8]{inputenc}

\newcommand{\norm}[1]{\left\lVert#1\right\rVert}
\newcommand{\sg}[1]{\operatorname{sg}\left(#1\right)}

\newcommand\target{\ensuremath{y}}
\newcommand{\encoder}{\operatorname{enc}}

\newcommand{\decoder}{\operatorname{dec}}
\newcommand{\discrete}[1]{z_d\left(#1\right)}

\renewcommand{\ae}{\operatorname{ae}}
\newcommand{\ad}{\operatorname{ad}}

\newcommand{\lp}{\operatorname{lp}}

\newcommand{\ci}{\mathrel{\text{\scalebox{1.07}{$\perp\mkern-10mu\perp$}}}}

\usepackage[accepted]{icml2018}

\begin{document}

\twocolumn[
\icmltitle{Fast Decoding in Sequence Models Using Discrete Latent Variables}




\begin{icmlauthorlist}
\icmlauthor{\L{}ukasz Kaiser}{g}
\icmlauthor{Aurko Roy}{g}
\icmlauthor{Ashish Vaswani}{g}
\icmlauthor{Niki Parmar}{g}
\icmlauthor{Samy Bengio}{g}
\icmlauthor{Jakob Uszkoreit}{g}
\icmlauthor{Noam Shazeer}{g}
\end{icmlauthorlist}

\icmlaffiliation{g}{Google Brain, Mountain View, California, USA}

\icmlcorrespondingauthor{\L{}ukasz Kaiser}{lukaszkaiser@google.com}
\icmlcorrespondingauthor{Aurko Roy}{aurkor@google.com}

\icmlkeywords{Machine Learning, Neural Machine Translation}

\vskip 0.3in
]



\printAffiliationsAndNotice{} 

\begin{abstract}
Autoregressive sequence models based on deep neural networks, such as
RNNs, Wavenet and the Transformer attain state-of-the-art results on many tasks.
However, they are difficult to parallelize and are thus slow at processing long sequences.
RNNs lack parallelism both during training and decoding, while 
architectures like WaveNet and Transformer are much more parallelizable
during training, yet still operate sequentially during decoding.
We present a method to extend sequence models using 
discrete latent variables that makes decoding much more parallelizable. 
We first auto-encode the target sequence into a shorter sequence of discrete latent variables,
which at inference time is generated autoregressively,
and finally decode the output sequence from this shorter 
latent sequence in parallel.
To this end, we introduce a novel method for constructing
a sequence of discrete latent variables and compare it with previously
introduced methods. Finally, we evaluate our model end-to-end
on the task of neural machine translation, 
where it is an order of magnitude faster at decoding than comparable
autoregressive models. While lower in BLEU than purely
autoregressive models, our model achieves higher scores than previously proposed
non-autoregressive translation models.
\end{abstract}

\section{Introduction}
Neural networks have been applied successfully
to a variety of tasks involving natural language.
In particular, recurrent neural networks (RNNs) with long
short-term memory (LSTM) cells \cite{hochreiter1997} in
a sequence-to-sequence configuration \cite{sutskever14} have
proven successful at tasks including machine translation
\cite{sutskever14,bahdanau2014neural,cho2014learning}, parsing
\cite{KVparse15}, and many others.
RNNs are inherently sequential, however, and thus tend to be slow
to execute on modern hardware optimized for parallel execution.
Recently, a number of more parallelizable sequence models were proposed
and architectures such as WaveNet \cite{wavenet}, ByteNet \cite{NalBytenet2017} and
the Transformer \cite{transformer} can indeed be trained faster due to improved parallelism.

When actually generating sequential output, however, their autoregressive nature 
still fundamentally prevents these models from taking full advantage of parallel computation.
When generating a sequence $y_1 \dots y_n$ in a canonical order, say from left to right, predicting
the symbol $y_t$ first requires generating all symbols $y_1 \dots y_{t-1}$ as the model predicts
\begin{align*}
    P(y_t|y_{t-1}\ y_{t-2}\dots\ y_1).
\end{align*}
During training, the ground truth is known so the conditioning on previous
symbols can be parallelized. But during decoding, this is a fundamental
limitation as at least $n$ sequential steps need to be made to generate
$y_1 \dots y_n$.

To overcome this limitation, we propose to introduce a sequence of
discrete latent variables $l_1 \dots l_m$, with $m < n$, that summarizes
the relevant information from the sequence $y_1 \dots y_n$.
We will still generate $l_1 \dots l_m$ autoregressively, but it will
be much faster as $m < n$ (in our experiments we mostly use $m = \frac{n}{8}$).
Then, we reconstruct each position in the sequence $y_1 \dots y_n$ from $l_1 \dots l_m$ in
parallel.

For the above strategy to work, we need to autoencode the target sequence
$y_1\dots y_n$ into a shorter sequence $l_1 \dots l_m$.
Autoencoders have a long history in deep learning \cite{reddim,deepbm,stackeddenoising,vae}.
Autoencoders mostly operate on continuous representations,
either by imposing a bottleneck \cite{reddim}, requiring them to remove added noise \cite{stackeddenoising},
or adding a variational component \cite{vae}. In our case though, we prefer
the sequence $l_1 \dots l_m$ to be discrete, as we use standard
autoregressive models to predict it. Despite some success \cite{textvae,textvae2}, predicting continuous latent representations
does not work as well as the discrete case in our setting.

However, using discrete latent variables can be challenging when
training models end-to-end. Three techniques recently have shown how
to successfuly use discrete variables in deep models: the Gumbel-Softmax
\cite{gs1,gs2}, VQ-VAE \cite{vqvae} and improved semantic
hashing \cite{isemhash}. We compare all these techniques in our setting
and introduce another one: decomposed vector quantization (DVQ) which
performs better than VQ-VAE for large latent alphabet sizes.

Using either DVQ or improved semantic hashing, we are able
to create a neural machine translation model that achieves good BLEU
scores on the standard benchmarks while being an order of magnitude faster
at decoding time than autoregressive models. A recent paper \cite{nonautoregnmt}
reported similar gain for neural machine translation. But their techniques are
hand-tuned for translation and require training with reinforcement learning.
Our latent variables are learned and the model is trained end-to-end,
so it can be applied to any sequence problem.
Despite being more generic, our model outperforms the hand-tuned technique
from \cite{nonautoregnmt} yielding better BLEU. To summarize, our main contributions are:
\begin{enumerate}[(1)]
    \item A method for fast decoding for autoregressive models.
    \item An improved discretization technique: the DVQ.
    \item The resulting Latent Transformer model, achieving good
      results on translation while decoding much faster.
\end{enumerate}

\section{Discretization Techniques}\label{sec:discretization}
In this section we introduce various \emph{discretization bottlenecks} used to 
train discrete autoencoders for the target sequence. 
We will use the notation from 
\cite{vqvae} where the target sequence \(\target\) is passed through an encoder,
\(\encoder\), to produce 
a continuous latent representation \(\encoder(\target) \in R^D\), where
\(D\) is the dimension of the latent space. Let \(K\) be the size of the 
discrete latent space and let \([K]\) denote the set $\{1, 2, \dots, K\}$.
The continuous latent
\(\encoder(\target)\) is subsequently passed through 
a discretization bottleneck to produce a discrete latent representation 
\(\discrete{\target} \in [K]\), and an input \(z_q(\target)\) to be passed to the decoder
\(\decoder\).
For integers \(i, m\) we will use  \(\tau_m(i)\) to 
denote the binary representation of 
\(i\) using \(m\) bits, with the inverse operation, i.e. conversion 
from binary to decimal denoted by \(\tau^{-1}_m\). 

\subsection{Gumbel-Softmax}
A discretization technique that has recently received a lot of interest is
the Gumbel-Softmax trick proposed by \cite{gs1, gs2}. 
In this case one simply projects the encoder output \(\encoder(\target)\)
using a learnable projection \(W \in R^{K \times D}\) to get the logits 
\(l = W\encoder(\target)\) with the discrete code \(\discrete{\target}\) being
defined as
\begin{align}
    \discrete{\target} = \arg\max_{i \in [K]} l_i.
\end{align}
The decoder input \(z_q(\target) \in R^D\) during evaluation and inference 
is computed using an embedding \(e \in R^{K \times D}\) 
where \(z_q(\target) = e_j\), where \(j = z_d(y)\). 
For training, the Gumbel-Softmax 
trick is used by generating samples 
\(g_1, g_2, \dots, g_K\) i.i.d samples 
from the Gumbel distribution:
\(g_i \sim -\log\left(-\log{u}\right)\), where \(u \sim \mathcal{U}(0, 1)\)
are uniform samples. Then as in \cite{gs1, gs2}, one computes the log-softmax
of \(l\) to get \(w \in R^K\):
\begin{align}
    w_i = \frac{\exp((l_i + g_i)/\tau)}{\sum_i \exp((l_i + g_i)/\tau)},
\end{align}
with the input to the decoder \(z_q(\target)\) being simply the matrix-vector product
\(we\).
Note that the Gumbel-Softmax trick makes the model differentiable
and thus it can be trained using backpropagation.

For low temperature \(\tau\) the vector \(w\)
is close to the 1-hot vector
representing the maximum index of \(l\), 
which is what is used during 
evaluation and testing. 
But at higher temperatures, 
it is an approximation (see Figure~1 in \citet{gs1}).

\subsection{Improved Semantic Hashing}\label{sec:semhash}
Another discretization technique proposed by \cite{isemhash} 
that has been recently explored 
stems from \emph{semantic hashing} \cite{semhash}. The main idea behind this
technique is to use a simple 
rounding bottleneck
after squashing the encoder
state \(z_e(\target)\) using a saturating sigmoid.
Recall the saturating sigmoid function from 
\cite{neural_gpu,extendedngpu}:
\begin{align} 
\sigma'(x) = \max(0, \min(1, 1.2\sigma(x) - 0.1)). 
\end{align}
During training, a Gaussian noise \(\eta \sim \mathcal{N}(0, 1)^{D}\)
is added to \(z_e(\target)\) which is then passed through a saturating sigmoid 
to get the vector \(f_e(\target)\):
\begin{align}
    f_e(\target) = \sigma'(z_e(\target) + \eta).
\end{align}
To compute the discrete latent representation, the binary 
vector \(g_e(\target)\) is constructed via rounding, i.e.:
\begin{align}
    g_e(\target)_i = \begin{cases} 1 \quad \text{if \(f_e(\target)_i > 0.5\)}\\
                        0 \quad \text{otherwise},
        \end{cases}
\end{align}
with the discrete latent code \(\discrete{\target}\) corresponding to 
\(\tau^{-1}_{\log{K}}(g(\target))\). 
The input to the decoder \(z_q(\target) \in R^D\) 
is computed using two embedding spaces 
\(e^1, e^2 \in R^{K \times D}\), with
\(z_q(\target) = e^1_{h_e(\target)} + e^2_{1 - h_e(\target)}\), 
where the function \(h_e\) is randomly chosen to be \(f_e\) or \(g_e\) 
half of the time during training, while \(h_e\) is set equal to \(g_e\) during inference.

\subsection{Vector Quantization}\label{sec:vqvae}
The Vector Quantized - Variational Autoencoder
(VQ-VAE) discretization bottleneck method was proposed in~\cite{vqvae}.
Note that vector quantization based methods have a long 
history of being used successfully in various Hidden Markov Model (HMM)
based machine learning models (see e.g., \cite{huang1989unified, lee1989sphinx}).
In VQ-VAE, the encoder output \(\encoder(\target) \in R^D\) is passed through 
a discretization bottleneck
using a nearest-neighbor lookup on embedding vectors \(e \in R^{K \times D}\).

More specifically, the decoder input \(z_q(\target)\) is defined as
\begin{align}
    z_q(\target) = e_k, \quad k = \arg\min_{j \in [K]}
    \norm{\encoder(\target) - e_j}_2.
\end{align}
The corresponding discrete latent \(\discrete{\target}\) 
is then the index \(k\) of the
embedding vector closest to \(\encoder(\target)\) in \(\ell_2\) distance. 
Let \(l_r\) be the reconstruction loss of the decoder \(\decoder\)
given \(z_q(\target)\), 
(e.g., the cross entropy loss); then the model is trained to minimize 
\begin{align}
    L = l_r + \beta \norm{\encoder(\target) - \sg{z_q(\target)}}_2,
\end{align}
where \(\sg{.}\) is the stop gradient operator defined as follows:
\begin{align}
    \sg{x} = \begin{cases} x \quad \text{forward pass} \\
                           0  \quad \text{backward pass}
             \end{cases}
\end{align}
We maintain an exponential moving average (EMA)
over the following two quantities: 1) the embeddings \(e_j\)
for every \(j \in [K]\) and, 2) the count \(c_j\)
measuring the number of encoder hidden states that 
have \(e_j\) as it's nearest neighbor. The counts are updated 
over a mini-batch of targets \(\{y_1, \dots y_l, \dots\}\) as:
\begin{align}\label{eq:ema1}
    c_j \gets \lambda c_j + (1 - \lambda) \sum_{l} {1}\left[z_q(\target_l) = e_j\right],
\end{align}
with the embedding \(e_j\) being subsequently updated as:
\begin{align}\label{eq:ema2}
    e_j \gets \lambda e_j + (1 - \lambda) \sum_{l} 
    \frac{{1}\left[z_q(\target_l) = e_j\right] \encoder(\target_l)}{c_j},
\end{align}
where \({1}[.]\) is the indicator function and \(\lambda\)
is a decay parameter which we set to \(0.999\) in our experiments.

\subsection{Decomposed Vector Quantization}\label{sec:dvq}
When the size of the discrete latent space \(K\) is large, 
then an issue with the approach of Section~\ref{sec:vqvae} is 
\emph{index collapse},
where only a few of the embedding vectors get trained
due to a rich getting richer phenomena. In particular, if an
embedding vector \(e_j\) is close to a 
lot of encoder outputs \(\encoder(\target_1), \dots,
\encoder(\target_i)\), then
it receives the strongest signal to get even closer via the EMA
update of Equations~\eqref{eq:ema1} and~\eqref{eq:ema2}. 
Thus only a few of the embedding vectors will end up actually being used. 
To circumvent this issue, we propose two variants 
of decomposing VQ-VAE that make more efficient use of the embedding
vectors for large values of \(K\). 

\subsubsection{Sliced Vector Quantization}\label{sec:sliced-vqvae}
The main idea behind this approach is to break up the encoder output 
\(\encoder(\target)\) into \(n_d\) smaller slices 
\begin{align}
\operatorname{enc}^1(\target) \circ \operatorname{enc}^2(\target) 
\cdots \circ \operatorname{enc}^{n_d}(\target),
\end{align}
where each \(\operatorname{enc}^i(\target)\) 
is a \(D/n_d\) dimensional vector and \(\circ\) denotes
the concatenation operation.  
Corresponding to each \(\operatorname{enc}^i(\target)\) we
have an embedding space \(e^i \in R^{K' \times D/{n_d}}\), where 
\(K' = 2^{\left(\log_2{K}\right) / n_d}\). 
Note that the reason for the particular choice of 
\(K'\) is information theoretic: using an embedding space of 
size \(K\) from Section~\ref{sec:vqvae} allows us to express 
discrete codes of size \(\log_2{K}\). In the case when we have 
\(n_d\) different slices, we want the total expressible size of the 
discrete code to be still \(\log_2{K}\) and so \(K'\) is 
set to \(2^{\left(\log_2{K}\right) / n_d}\).
We now compute nearest neighbors
for each subspace as:
\begin{align}
    z^i_q(\target) = e^i_{k_i}, \quad k_i = \arg\min_{j \in [K]}
    \norm{\operatorname{enc}^i(\target) - e^i_j}_2,
\end{align}
with the decoder input being \(z_q(\target) = z^1_q(\target) \circ \cdots 
\circ z^{n_d}_q(\target)\).

The training objective \(L\) is the same as in Section~\ref{sec:vqvae}, 
with each embedding space \(e^i\) trained individually via EMA updates
from \(\operatorname{enc}^i(\target)\) over a mini-batch of targets
\(\{\target_1, \dots, \target_l, \dots\}\):
\begin{align}
    c^i_j &\gets \lambda c^i_j + (1 - \lambda) \sum_l {1}\left[z_q(\target_l) = e^i_j\right]\\
    e^i_j &\gets \lambda e^i_j + (1 - \lambda) \frac{\sum_{l} {1}\left[z^i_q(\target_l) = e^i_j\right] 
    \encoder^i(\target_l)}{c^i_j},
\end{align}
where \({1}[.]\) is the indicator function as before, and \(\lambda\)
is the decay parameter.

Then the discrete latent code \(\discrete{\target}\) 
is now defined as
\begin{align}
\discrete{\target} = \tau^{-1}_{\log_2{K}}\left(\tau_{\log_2{K'}}(k_1) \circ 
\cdots \circ \tau_{\log_2{K'}}(k_{n_d}) \right).
\end{align}
Observe that when \(n_d = 1\), the sliced Vector Quantization reduces
to the VQ-VAE of \cite{vqvae}. On the other hand, when 
\(n_d = \log_2{K}\), sliced DVQ is equivalent to improved semantic 
hashing of Section~\ref{sec:semhash} loosely speaking: the individual
table size \(K'\) for each slice \(\operatorname{enc}^i(\target)\) 
is \(2\), and it gets rounded to \(0\) or \(1\) depending on which 
embedding is closer. However, the rounding bottleneck in semantic hashing
of Section~\ref{sec:semhash} proceeds via a saturating sigmoid and 
thus strictly speaking, the two techniques are different.

Note that similar decomposition approaches to 
vector quantization in the context of HMMs have been studied in the past under the name
\emph{multiple code-books}, see for instance \cite{huang1989multiple, rogina1994learning, peinado1996discriminative}.
The approach of sliced Vector Quantization has also been studied more recently in the context of
clustering, under the name of Product or Cartesian Quantization 
in \cite{jegou2011product, norouzi2013cartesian}. A more recent work 
\cite{shu2018compressing} explores a similar quantization 
approach coupled with the Gumbel-Softmax trick to learn compressed
word embeddings (see also \cite{word2bits}).

\subsubsection{Projected Vector Quantization}\label{sec:projected-vqvae}
Another natural way to decompose Vector Quantization is to use a set of 
fixed randomly initialized projections 
\begin{align}
\left\{\pi^i \in R^{D \times D/{n_d}} \mid i \in [n_d]\right\} 
\end{align}
to project the encoder output \(\encoder(\target)\) into a
\(R^{D/n_d}\)-dimensional subspace. 
For \(\operatorname{enc}^i(\target) = \pi^i(\operatorname{\target}) \in R^{D/n_d}\)
we have an embedding
space \(e^i \in R^{K' \times D/n_d}\), where \(K' = 2^{\left(\log_2{K}\right) / n_d}\)
as before. The training objective, embeddings update,
the input \(z_q(\target)\) to the decoder,
and the discrete latent representation \(\discrete{\target}\)
is computed exactly as in Section~\ref{sec:sliced-vqvae}. Note that when 
\(n_d = 1\), projected Vector Quantization reduces to the VQ-VAE of 
\cite{vqvae} with an extra encoder layer corresponding to the projections 
\(\pi^i\). 
Similarly, when 
\(n_d = \log_2{K}\), projected DVQ is equivalent to improved semantic 
hashing of Section~\ref{sec:semhash} with the same analogy 
as in Section~\ref{sec:sliced-vqvae}, except the encoder now has 
an extra layer. The VQ-VAE paper \cite{vqvae} also use multiple latents
in the experiments reported on CIFAR-10 and in Figure 5,
using an approach
similar to what we call projected DVQ.

\section{Latent Transformer}\label{sec:lt}
Using the discretization techniques from Section~\ref{sec:discretization}
we can now introduce the Latent Transformer (LT) model.
Given an input-output pair $(x,y) = (x_1, \dots x_k, y_1, \dots y_n)$
the LT will make use of the following components.
\begin{itemize}
\item The function $\ae(y,x)$ will autoencode $y$ into a shorter sequence $l = l_1, \dots, l_m$ of discrete latent variables
  using the discretization bottleneck from Section~\ref{sec:discretization}.
\item The latent prediction model $\lp(x)$  (a Transformer) will autoregressively predict $l$ based on $x$.
\item The decoder $\ad(l, x)$ is a parallel model that will decode $y$ from $l$ and the input sequence $x$.
\end{itemize}

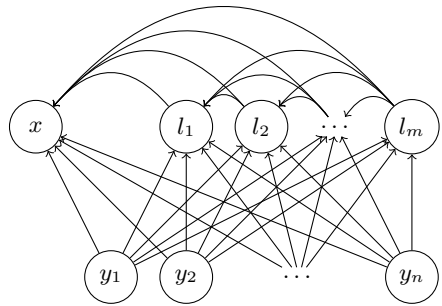
\begin{figure}
\begin{center}
\begin{tikzpicture}[yscale=2.0,xscale=1.0] \footnotesize

\node[draw,circle,minimum size=0.7cm] (x)  at (-2.0, 1.0) {$x$};
\node[draw,circle,minimum size=0.7cm] (l1) at (0.0, 1.0) {$l_1$};
\node[draw,circle,minimum size=0.7cm] (l2) at (1.0, 1.0) {$l_2$};
\node (li) at (2.0, 1.0) {$\dots$};
\node[draw,circle,minimum size=0.7cm] (lm) at (3.0, 1.0) {$l_m$};

\path[->] (l1) edge[bend right] (x);
\path[->] (l2) edge[bend right] (x);
\path[->] (l2) edge[bend right] (l1);
\path[->] (li) edge[bend right] (x);
\path[->] (li) edge[bend right] (l1);
\path[->] (li) edge[bend right] (l2);
\path[->] (lm) edge[bend right] (x);
\path[->] (lm) edge[bend right] (l1);
\path[->] (lm) edge[bend right] (l2);
\path[->] (lm) edge[bend right] (li);

\node[draw,circle,minimum size=0.7cm] (y1) at (-1.0, 0.0) {$y_1$};
\node[draw,circle,minimum size=0.7cm] (y2) at (0.0, 0.0) {$y_2$};
\node (yi) at (1.5, 0.0) {$\dots$};
\node[draw,circle,minimum size=0.7cm] (yn) at (3.0, 0.0) {$y_n$};

\draw[->] (y1) -- (x);
\draw[->] (y1) -- (l1);
\draw[->] (y1) -- (l2);
\draw[->] (y1) -- (li);
\draw[->] (y1) -- (lm);

\draw[->] (y2) -- (x);
\draw[->] (y2) -- (l1);
\draw[->] (y2) -- (l2);
\draw[->] (y2) -- (li);
\draw[->] (y2) -- (lm);

\draw[->] (yi) -- (x);
\draw[->] (yi) -- (l1);
\draw[->] (yi) -- (l2);
\draw[->] (yi) -- (li);
\draw[->] (yi) -- (lm);

\draw[->] (yn) -- (x);
\draw[->] (yn) -- (l1);
\draw[->] (yn) -- (l2);
\draw[->] (yn) -- (li);
\draw[->] (yn) -- (lm);

\end{tikzpicture}
\end{center}
\caption{Dependence structure of the Latent Transformer in the form of a graphical model.
  We merged all inputs $x_1 \dots x_k$ into a single node for easier visibility and we
  draw an arrow from node $a$ to $b$ if the probability of $a$ depends on the generated $b$.}
\label{fig:dep}
\end{figure}

The functions $\ae(y,x)$ and $\ad(l,x)$ together form an autoencoder
of the targets $\target$ that has additional access to the input sequence $x$.
For the autoregressive latent prediction
we use a Transformer \cite{transformer}, a model based on multiple self-attention layers
that was originally introduced in the context of neural machine translation. In this work
we focused on the autoencoding functions and did not tune the Transformer: we used all
the defaults from the baseline provided by the Transformer authors ($6$ layers, hidden size
of $512$ and filter size of $4096$) and only varied parameters relevant to $\ae$ and $\ad$,
which we describe below.
The three components above give rise to two losses:
\begin{itemize}
\item The autoencoder reconstruction loss $l_{r}$ coming from comparing $\ad(\ae(y, x), x)$ to $y$.
\item The latent prediction loss $l_{lp}$ that comes from comparing $l = \ae(y,x)$ to the generated $\lp(x)$.
\end{itemize}
We train the LT model by minimizing $l_{r} + l_{lp}$.
Note that the final outputs $y$ are generated only depending on
the latents $l$ but not on each other, as depicted in Figure~\ref{fig:dep}.
In an autoregressive model, each $y_i$ would have a dependence on
all previous $y_j, j < i$, as is the case for $l$s in Figure~\ref{fig:dep}.

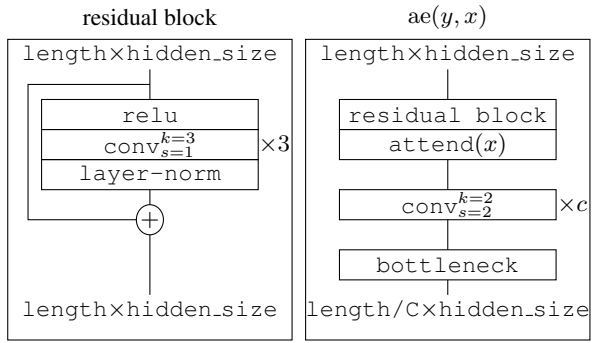
\begin{figure}[h]
\begin{center}
\begin{tikzpicture}[yscale=2.0,xscale=1.8] \footnotesize

\node at (1.0, 2.15) {residual block};
\draw (-0.05, 0.0) rectangle (2.05, 2.0);

\node at (1.0, 1.9) {\texttt{length$\times$hidden\_size}};
\draw (1.0, 1.8) -- (1.0, 1.6);
\draw (0.2, 1.6) rectangle (1.8, 1.4);
\node at (1.0, 1.5) {\texttt{relu}};
\draw (0.2, 1.4) rectangle (1.8, 1.2);
\node at (1.0, 1.3) {\texttt{conv}$^{k=3}_{s=1}$};
\node at (1.92, 1.3) {$\times 3$};
\draw (0.2, 1.2) rectangle (1.8, 1.0);
\node at (1.0, 1.1) {\texttt{layer-norm}};
\draw (1.0, 1.0) -- (1.0, 0.9);
\node at (1.0, 0.8) {$+$};
\draw (1.0, 0.8) circle (0.1cm);
\draw (1.0, 1.7) -- (0.1, 1.7) -- (0.1, 0.8) -- (0.9, 0.8);
\draw (1.0, 0.7) -- (1.0, 0.3);
\node at (1.0, 0.2) {\texttt{length$\times$hidden\_size}};

\begin{scope}[xshift=2.2cm]
\node at (1.0, 2.15) {$\ae(y,x)$};
\draw (-0.05, 0.0) rectangle (2.05, 2.0);

\node at (1.0, 1.9) {\texttt{length$\times$hidden\_size}};
\draw (1.0, 1.8) -- (1.0, 1.6);
\draw (0.2, 1.6) rectangle (1.8, 1.4);
\node at (1.0, 1.5) {\texttt{residual block}};
\draw (0.2, 1.4) rectangle (1.8, 1.2);
\node at (1.0, 1.3) {\texttt{attend}$(x)$};
\draw (1.0, 1.2) -- (1.0, 1.0);
\draw (0.2, 1.0) rectangle (1.8, 0.8);
\node at (1.0, 0.9) {\texttt{conv}$^{k=2}_{s=2}$};
\node at (1.92, 0.9) {$\times c$};
\draw (1.0, 0.8) -- (1.0, 0.6);
\draw (0.2, 0.6) rectangle (1.8, 0.4);
\node at (1.0, 0.5) {\texttt{bottleneck}};
\draw (1.0, 0.4) -- (1.0, 0.3);
\node at (1.0, 0.2) {\texttt{length/C$\times$hidden\_size}};
\end{scope}

\end{tikzpicture}
\end{center}
\caption{Architecture of $\ae(y,x)$. We write \texttt{conv}$^{k=a}_{s=b}$ to
  denote a 1D convolution layer with kernel size $a$ and stride $b$.}
\label{fig:ae}
\end{figure}

\paragraph{The function $\ae(y,x)$.}
The autoencoding function $\ae(y,x)$ we use is a stack
of residual convolutions followed by an attention layer
attending to $x$ and a stack of strided convolutions.
We first apply to $y$ a $3$-layer block of
$1$-dimensional convolutions with kernel size $3$ and padding with $0$s on both
sides (\texttt{SAME}-padding). We use ReLU non-linearities between the layers and
layer-normalization \cite{layernorm2016}. Then, we add the input to the result, forming a residual block.
Next we have an encoder-decoder attention layer with dot-product attention,
same as in \cite{transformer}, with a residual connection.
Finally, we process the result with a convolution with kernel size $2$ and stride $2$,
effectively halving the size of $s$. We do this strided processing $c$ times
so as to decrease the length $C = 2^c$ times (later $C = \frac{n}{m}$).
The result is put through the discretization bottleneck of Section~\ref{sec:discretization}.
The final compression function is given by $\ae(y,x) = z_q(\target)$ and
the architecture described above is depicted in Figure~\ref{fig:ae}.

\paragraph{The function $\ad(l,x)$.}
To decode from the latent sequence $l = \ae(y,x)$,
we use the function $\ad(l,x)$. It consists of $c$
steps that include up-convolutions that double the length,
so effectively it increases the length $2^c = C = \frac{n}{m}$ times.
Each step starts with the residual block, followed by
an encoder-decoder attention to $x$ (both as in
the ae function above). Then it applies an up-convolution,
which in our case is a feed-forward layer (equivalently a kernel-$1$ 
convolution) that doubles the internal dimension, followed
by a reshape to twice the length. The result after the $c$
steps is then passed to a self-attention decoder, same as
in the Tranformer model \cite{transformer}.

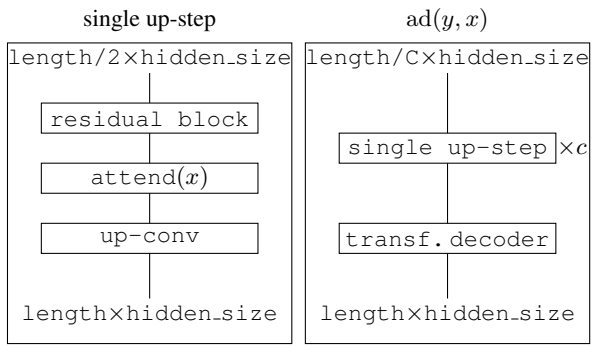
\begin{figure}
\begin{center}
\begin{tikzpicture}[yscale=2.0,xscale=1.8] \footnotesize

\node at (1.0, 2.15) {single up-step};
\draw (-0.05, 0.0) rectangle (2.05, 2.0);

\node at (1.0, 1.9) {\texttt{length/2$\times$hidden\_size}};
\draw (1.0, 1.8) -- (1.0, 1.6);
\draw (0.2, 1.6) rectangle (1.8, 1.4);
\node at (1.0, 1.5) {\texttt{residual block}};
\draw (1.0, 1.4) -- (1.0, 1.2);
\draw (0.2, 1.2) rectangle (1.8, 1.0);
\node at (1.0, 1.1) {\texttt{attend}$(x)$};
\draw (1.0, 1.0) -- (1.0, 0.8);
\draw (0.2, 0.8) rectangle (1.8, 0.6);
\node at (1.0, 0.7) {\texttt{up-conv}};
\draw (1.0, 0.6) -- (1.0, 0.3);
\node at (1.0, 0.2) {\texttt{length$\times$hidden\_size}};

\begin{scope}[xshift=2.2cm]
\node at (1.0, 2.15) {$\ad(y,x)$};
\draw (-0.05, 0.0) rectangle (2.05, 2.0);

\node at (1.0, 1.9) {\texttt{length/C$\times$hidden\_size}};
\draw (1.0, 1.8) -- (1.0, 1.4);
\draw (0.2, 1.2) rectangle (1.8, 1.4);
\node at (1.0, 1.3) {\texttt{single up-step}};
\node at (1.92, 1.3) {$\times c$};
\draw (1.0, 1.2) -- (1.0, 0.8);
\draw (0.2, 0.8) rectangle (1.8, 0.6);
\node at (1.0, 0.7) {\texttt{transf.$\,$decoder}};
\draw (1.0, 0.6) -- (1.0, 0.3);
\node at (1.0, 0.2) {\texttt{length$\times$hidden\_size}};
\end{scope}

\end{tikzpicture}
\end{center}
\caption{Architecture of $\ad(l,x)$. We write \texttt{upconv} to
  denote a 1D up-convolution layer.}
\label{fig:ad}
\end{figure}

Note that at the beginning of training (for the first 10K steps),
we give the true targets $y$ to the transformer-decoder here, instead of the decompressed
latents $l$. This pre-training ensures that the self-attention part has reasonable gradients
that are then back-propagated to the convolution stack and then back to the $\ae$ function
and the discretization bottleneck of Section~\ref{sec:discretization}.

\section{Related Work}

\paragraph{Neural Machine Translation.}
Machine translation using deep neural networks achieved great success with
sequence-to-sequence models \cite{sutskever14,bahdanau2014neural,cho2014learning}
that used recurrent neural networks (RNNs) with LSTM cells \cite{hochreiter1997}.
The basic sequence-to-sequence architecture is composed of an RNN encoder which reads
the source sentence one token at a time and transforms it into a fixed-sized state vector.
This is followed by an RNN decoder, which generates the target sentence,
one token at a time, from the state vector.
While a pure sequence-to-sequence recurrent neural network can already
obtain good translation results \cite{sutskever14,cho2014learning},
it suffers from the fact that the whole input sentence
needs to be encoded into a single fixed-size vector. This clearly
manifests itself in the degradation of translation quality
on longer sentences and was overcome in \cite{bahdanau2014neural}
by using a neural model of attention. 
Convolutional architectures have been used to obtain good results in word-level
neural machine translation starting from \cite{KalchbrennerB13} and later in
\cite{MengLWLJL15}. These early models used a standard RNN on top of
the convolution to generate the output, which creates a bottleneck and hurts performance.
Fully convolutional neural machine translation without this bottleneck
was first achieved in \cite{extendedngpu} and \cite{NalBytenet2017}.
The model in \cite{extendedngpu} (Extended Neural GPU) used a recurrent stack
of gated convolutional layers, while the model in \cite{NalBytenet2017} (ByteNet)
did away with recursion and used left-padded convolutions in the decoder.
This idea, introduced in WaveNet \cite{wavenet}, significantly improves
efficiency of the model. The same technique was improved
in a number of neural translation models recently, including \cite{JonasFaceNet2017}
and \cite{slicenet}. Instead of convolutions, one can use stacked
self-attention layers. This was introduced in the Transformer model 
\cite{transformer} and has significantly improved state-of-the-art
in machine translation while also improving the speed of training.
Thus, we use the Transformer model as a baseline in this work.

\paragraph{Autoencoders and discretization bottlenecks.} 
Variational autoencoders were first introduced in \cite{vae, rezende2014stochastic}, 
however training them for discrete latent variable models has been challenging. 
The NVIL estimator of \cite{mnih2014neural} proposes using a single sample objective
to optimize the variational lower bound, while VIMCO \cite{mnih2016variational} 
proposes using a muliti-sample objective of \cite{burda2015importance} which further
speeds up convergence by using multiple samples from the inference network.
There have also been several discretization bottlenecks proposed recently
that have been used successfully in various learning tasks,
see Section~\ref{sec:discretization} for a more detailed 
description of the techniques directly relevant to this work.
Other recent works with similar approach to autoencoding include \cite{aepredict}.

\paragraph{Non-autoregressive Neural Machine Translation.}
Much of the recent state of the art models in Neural Machine Translation
are auto-regressive, meaning that the model consumes previously generated
tokens to predict the next one. A recent work that attempts to speed up
decoding by training a non-autotregressive model is \cite{nonautoregnmt}.
The approach of \cite{nonautoregnmt} is to use the self-attention Transformer
model of \cite{transformer}, together with the REINFORCE algorithm~\cite{williams1992simple}
to model the \emph{fertilities} of words to tackle the multi-modality problem in 
translation. However, the main drawback of this work is the need for extensive
fine-tuning to make policy gradients work for REINFORCE, 
as well as the issue that this approach only works for machine translation
and is not generic, so it cannot be directly applied to other sequence learning tasks.

\paragraph{Graphical models.} The core of our approach to fast decoding consists
of finding a sequence $l$ of latent variables such that we can predict the output sequence $y$
in parallel from $l$ and the input $x$. In other words, we assume that each token $y_i$ is
conditionally independent of all other tokens $y_j$ ($j \neq i$) given $l$ and $x$:
$y_i \ci y_j\ |\ l, x$. Our autoencoder is thus learning to create a one-layer
graphical model with $m$ variables ($l_1 \dots l_m$) that can then be used
to predict $y_1 \dots y_n$ independently of each other.

\section{Experiments}
We train the Latent Transformer with the base configuration
to make it comparable to both the autoregressive baseline \cite{transformer}
and to the recent non-autoregressive NMT results \cite{nonautoregnmt}.
We used around 33K subword units as vocabulary and implemented our model
in TensorFlow \cite{tensorflow}. Our implementation, together with
hyper-parameters and everything needed to reproduce our results is
available as open-source\footnote{The code is available under \url{redacted}.}.

For non-autoregressive models, it is beneficial to generate a number
of possible translations and re-score them with an autoregressive model.
This can be done in parallel, so it is still fast, and it improves performance.
This is called \emph{noisy parallel decoding} in \cite{nonautoregnmt} and we
include results both with and without it. The best BLEU scores obtained by
different methods are summarized in Table~\ref{tab:res}. As you can see,
our method with re-scoring almost matches the baseline autoregressive
model without beam search.

\begin{table}
\begin{center}
\begin{tabular}{|l||c|}
\hline
{\bf Model}                             & {\bf BLEU} \\ \hline
Baseline Transformer [1]                  & 27.3  \\
Baseline Transformer [2]                  & 23.5 \\
Baseline Transformer [2] (no beam-search) & 22.7 \\
\hline
NAT+FT (no NPD) [2]                     & 17.7 \\
LT without rescoring $\left(\frac{n}{m}=8\right)$  & 19.8 \\
\hline
NAT+FT (NPD rescoring 10) [2]           & 18.7 \\
LT rescornig top-10  $\left(\frac{n}{m}=8\right)$  & 21.0 \\
\hline
NAT+FT (NPD rescoring 100) [2]          & 19.2 \\
LT rescornig top-100 $\left(\frac{n}{m}=8\right)$  & 22.5 \\
\hline
\end{tabular}
\end{center}
\caption{BLEU scores (the higher the better) on the WMT English-German
  translation task on the \texttt{newstest2014} test set.
  The acronym NAT corresponds to the Non-Autoregressive Transformer,
  while FT denotes an additional Fertility Training and NPD denotes Nosiy Parallel
  Decoding, all of them from \cite{nonautoregnmt}. The acronym LT denotes the Latent 
  Transformer from Section~\ref{sec:lt}.
  Results reported for LT are from this work, 
  the others are from \cite{nonautoregnmt} [2] except for the first
  baseline Transformer result which is from \cite{transformer} [1].}
\label{tab:res}
\end{table}

To get a better understanding of the non-autoregressive models, we focus
on performance without rescoring and investigate different variants of
the Latent Transformer. We include different discretization bottlenecks,
and report the final BLEU scores together with decoding speeds
in Table~\ref{tab:ablations}. The LT is slower in non-batch mode
than the simple NAT baseline of \cite{nonautoregnmt}, which might
be caused by system differences (our code is in TensorFlow
and has not been optimized, while their implementation is in Torch).
Latency at higher batch-size is much smaller, showing that the speed
of the LT can still be significantly improved with batching.
The choice of the discretization bottleneck seems to have a small
impact on speed and both DVQ and improved semantic hashing yield
good BLEU scores, while VQ-VAE fails in this context (see below
for a discussion).

\begin{table}
\begin{center}
\begin{tabular}{|l||c|c|c|c|}
\hline
{\bf Model}   & BLEU & \multicolumn{2}{|c|}{\textbf{Latency}} \\ \cline{3-4}
& & \(b = 1\) & \(b = 64\) \\
\hline 
Baseline (no beam-search) & 22.7 & 408 ms  & - \\
NAT                       & 17.7 & 39 ms & - \\
NAT+NPD=10                & 18.7 & 79 ms & - \\
NAT+NPD=100               & 19.2 & 257 ms & - \\
\hline
LT, Improved Semhash & 19.8 & 105 ms & 8 ms\\
LT, VQ-VAE & 2.78 & 148 ms & 7 ms\\
LT, s-DVQ  & 19.7 & 177 ms & 7 ms\\
LT, p-DVQ & 19.8 & 182 ms & 8 ms\\
\hline
\end{tabular}
\end{center}
\caption{BLEU scores and decode times on the WMT English-German translation task
  on the \texttt{newstest2014} test set for different variants of the LT
  with \(\frac{n}{m} = 8\) and \(D = 512\) and \(n_d = 2\) with s-DVQ and p-DVQ representing 
  sliced and projected DVQ respectively. 
  The LT model using improved semantic hashing from Section~\ref{sec:semhash}
  uses \(\log_2{K} = 14\), while the one using VQ-VAE and DVQ from Sections~\ref{sec:vqvae}
  and \ref{sec:dvq}
  uses \(\log_2{K} = 16\).
  For comparison, we include the baselines from \cite{nonautoregnmt}. 
  We report the time to decode per sentence averaged over the 
  whole test set as in \cite{nonautoregnmt}; decoding is implemented 
  in Tensorflow on a Nvidia GeForce GTX 1080. The batch size used during decoding
  is denoted by \(b\) and we report both \(b=1\) and \(b=64\).
  }
\label{tab:ablations}
\end{table}

\section{Discussion} \label{sec:discussion}
Since the discretization bottleneck is critical to obtaining
good results for fast decoding of sequence models, we 
focused on looking into it, especially in conjunction with 
the size \(K\) of the latent vocabulary, the dimension \(D\)
of the latent space, and the number of decompositions \(n_d\)
for DVQ.

\begin{figure}[!htb]
    \centering
    \includegraphics[scale=.20]{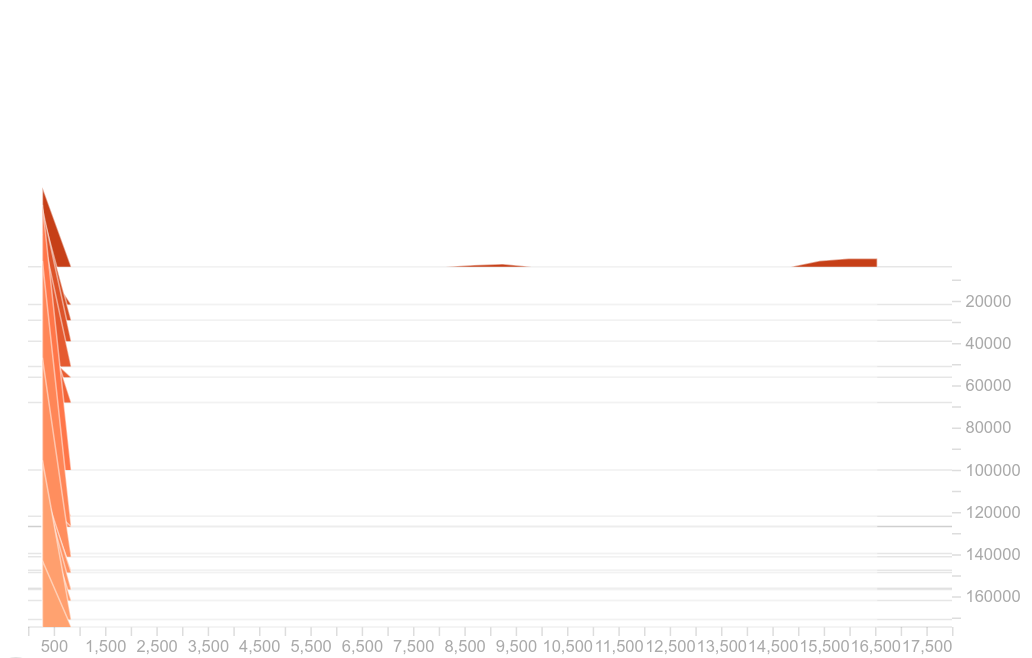}
    \caption{Histogram of discrete latent usage for VQ-VAE from \cite{vqvae}, or equivalently sliced 
    DVQ with \(n_d = 1\) and \(\log_2{K} = 14\).  
    The \(x\)-axis corresponds to the different possible discrete
    latents, while the \(y\)-axis corresponds to the progression of 
    training steps (time increases in a downwards direction). 
    Notice \emph{index collapse} in the 
    vanilla VQ-VAE where only a few latents ever get used.}
    \label{fig:vqvae-14-1}
\end{figure}

\begin{figure}[!htb]
    \centering
    \includegraphics[scale=.20]{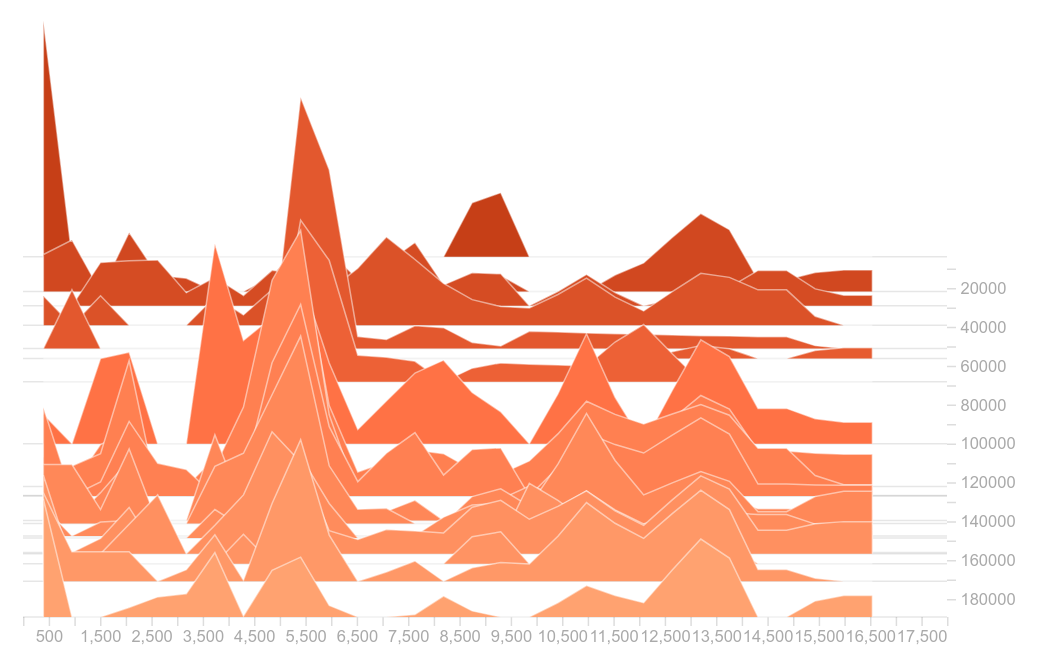}
    \caption{Histogram of discrete latent usage for sliced 
    DVQ with \(n_d = 2\) and \(\log_2{K} = 14\). 
    The \(x\)-axis corresponds to the different possible discrete
    latents, while the \(y\)-axis corresponds to the progression of training steps
    (time increases in a downwards direction). 
    Notice the diversity of latents used in this case.}
    \label{fig:vqvae-14-2}
\end{figure}

An issue with the VQ-VAE of \cite{vqvae} that motivated 
the introduction of DVQ in Section~\ref{sec:vqvae} is 
\emph{index collapse}, where only a few embeddings are 
used and subsequently trained. This can be visualized in 
the histogram of Figure~\ref{fig:vqvae-14-1}, where the 
\(x\)-axis corresponds to the possible values of the 
discrete latents (in this case \(\{1, \dots, K\}\)),
and the \(y\)-axis corresponds to the training progression of
the model (time steps increase in a downward direction). 
On the other hand, using the DVQ from Section~\ref{sec:sliced-vqvae}
with \(n_d = 2\) leads to a much more balanced use of the available
discrete latent space, as can be seen from Figure~\ref{fig:vqvae-14-2}.
We also report the percentage of available latent code-words used
for different settings of \(n_d\) in Table~\ref{tab:percentage}; 
the usage of the code-words is maximized for \(n_d = 2\). 

The other variables for DVQ are the choice of the decomposition, and
the number \(n_d\) of decompositions. For the projected DVQ, we use fixed
projections \(\pi_i\)'s initialized using the Glorot initializer 
\citep{glorot2010understanding}.
We also found that the optimal number of decompositions for 
our choice of latent vocabulary size \(\log_2{K} = 14\) and \(16\)
was \(n_d = 2\), with \(n_d = 1\) (i.e., regular VQ-VAE) performing
noticeably worse (see~Table~\ref{tab:ablations} and Figure~\ref{fig:vqvae-14-1}).
Setting higher values of \(n_d\) led to a decline in performance, possibly 
because the expressive power (\(\log_2{K'}\)) was reduced for each 
decomposition, and the model also ended up using fewer latents
(see Table~\ref{tab:percentage}).

\begin{table}
\begin{center}
\begin{tabular}{|l||c|c|c|c|}
\hline
{\bf \(n_d\)}    & {\bf Percentage of latents used} \\ \hline
1  & \(5\%\) \\
2  & \(74.5\%\) \\
4  & \(15.6\%\) \\
8 & \(31.2\%\) \\ 
\hline
\end{tabular}
\end{center}
\caption{Percentage of latent codewords used by the Decomposed Vector Quantization (DVQ)
of Section~\ref{sec:sliced-vqvae} for \(\log_2{K} = 16\) and \(D = 512\) after \(500,000\) steps.
Note that when \(n_d = 1\), i.e. for vanilla VQ-VAE, only \(5\%\) of the available \(2^{16}\) discrete 
latents (roughly \(3000\)) are used. The latent usage is maximized 
for \(n_d = 2\).}
\label{tab:percentage}
\end{table}

Another important point about LT is that it allows making different trade-offs
by tuning the $\frac{n}{m}$ fraction of the length of the original
output sequence to the length of the latent sequence. As $\frac{n}{m}$ increases,
so does the parallelism and decoding speed, but the latents need to encode more
and more information to be able to decode the outputs in parallel.
To study this tradeoff, we measure the reconstruction loss (the perplexity of
the reconstructed $y$ vs the original) for different $\frac{n}{m}$ and
varying the number of bits in the latent variables. The results,
presented in Table~\ref{tab:tradeoff}, show clearly that reconstruction
get better, as expected, if the latent state has more bits or is used
to compress less subword units.

\begin{table}
\begin{center}
\begin{tabular}{|l||c|c|}
\hline
$\frac{n}{m}$ & $\log_{2}K = 8$ & $\log_{2}K = 14$ \\ \hline
 $2$ & 1.33 & 0.64 \\
 $4$ & 2.04 & 1.26 \\
 $8$ & 2.44 & 1.77 \\
\hline
\end{tabular}
\end{center}
\caption{Log-perplexities of autoencoder reconstructions on the development set (\texttt{newstest2013})
  for different values of $\frac{n}{m}$ and numbers of bits in
  latent variables (LT trained for 250K steps).}
\label{tab:tradeoff}
\end{table}

\section{Conclusions}

Autoregressive sequence models based on deep neural networks were made
successful due to their applications in machine translation \cite{sutskever14}
and have since yielded state-of-the-art results on a number of tasks.
With models like WaveNet and Transformer, it is possible to train them
fast in a parallel way, which opened the way to applications to longer sequences,
such as WaveNet for sound generation \cite{wavenet} or Transformer for
long text summarization \cite{wikipedia} and image generation \cite{imagetrans}.
The key problem appearing in these new applications is the slowness of decoding:
it is not practical to wait for minutes to generate a single example.
In this work, while still focusing on the original problem of machine translation,
we lay the groundwork for fast decoding for sequence models in general.
While the latent transformer does not yet recover the full performance
of the autoregressive model, it is already an order of magnitude faster
and performs better than a heavily hand-tuned, task-specific non-autoregressive model.

In the future, we plan to improve both the speed and the accuracy of
the latent transformer. A simple way to improve speed that we did not
yet try is to use the methods from this work in a hierarchical way.
As illustrated in Figure~\ref{fig:dep}, the latents are still generated
autoregressively which takes most of the time for longer sentences.
In the future, we will apply the LT model to generate the latents
in a hierarchical manner, which should result in further speedup.
To improve the BLEU scores, on the other hand, we intend to
investigate methods related to Gibbs sampling or even make
the model partially autoregressive. For example, one could generate
only the odd-indexed outputs, $y_1 y_3 y_5 \dots$, based on
the latent symbols $l$, and then generate the even-indexed ones
based on both the latents and the odd-indexed outputs.
We believe that including such techniques has the potential to remove
the gap between fast-decoding models and purely autoregressive ones
and will lead to many new applications.
\bibliography{deeplearn}

\begin{thebibliography}{48}
\providecommand{\natexlab}[1]{#1}
\providecommand{\url}[1]{\texttt{#1}}
\expandafter\ifx\csname urlstyle\endcsname\relax
  \providecommand{\doi}[1]{doi: #1}\else
  \providecommand{\doi}{doi: \begingroup \urlstyle{rm}\Url}\fi

\bibitem[Abadi et~al.(2015)Abadi, Agarwal, Barham, Brevdo, Chen, Citro,
  Corrado, Davis, Dean, Devin, Ghemawat, Goodfellow, Harp, Irving, Isard, Jia,
  Jozefowicz, Kaiser, Kudlur, Levenberg, Mané, Monga, Moore, Murray, Olah,
  Schuster, Shlens, Steiner, Sutskever, Talwar, Tucker, Vanhoucke, Vasudevan,
  Viégas, Vinyals, Warden, Wattenberg, Wicke, Yu, and Zheng]{tensorflow}
Abadi, Martín, Agarwal, Ashish, Barham, Paul, Brevdo, Eugene, Chen, Zhifeng,
  Citro, Craig, Corrado, Greg, Davis, Andy, Dean, Jeffrey, Devin, Matthieu,
  Ghemawat, Sanjay, Goodfellow, Ian, Harp, Andrew, Irving, Geoffrey, Isard,
  Michael, Jia, Yangqing, Jozefowicz, Rafal, Kaiser, Lukasz, Kudlur, Manjunath,
  Levenberg, Josh, Mané, Dan, Monga, Rajat, Moore, Sherry, Murray, Derek,
  Olah, Chris, Schuster, Mike, Shlens, Jonathon, Steiner, Benoit, Sutskever,
  Ilya, Talwar, Kunal, Tucker, Paul, Vanhoucke, Vincent, Vasudevan, Vijay,
  Viégas, Fernanda, Vinyals, Oriol, Warden, Pete, Wattenberg, Martin, Wicke,
  Martin, Yu, Yuan, and Zheng, Xiaoqiang.
\newblock Tensorflow: Large-scale machine learning on heterogeneous distributed
  systems, 2015.
\newblock URL \url{http://download.tensorflow.org/paper/whitepaper2015.pdf}.

\bibitem[Ba et~al.(2016)Ba, Kiros, and Hinton]{layernorm2016}
Ba, Jimmy~Lei, Kiros, Jamie~Ryan, and Hinton, Geoffrey~E.
\newblock Layer normalization.
\newblock \emph{arXiv preprint arXiv:1607.06450}, 2016.
\newblock URL \url{http://arxiv.org/abs/1607.06450}.

\bibitem[Bahdanau et~al.(2014)Bahdanau, Cho, and Bengio]{bahdanau2014neural}
Bahdanau, Dzmitry, Cho, Kyunghyun, and Bengio, Yoshua.
\newblock Neural machine translation by jointly learning to align and
  translate.
\newblock \emph{CoRR}, abs/1409.0473, 2014.
\newblock URL \url{http://arxiv.org/abs/1409.0473}.

\bibitem[Bowman et~al.(2016)Bowman, Vilnis, Vinyals, Dai, J{\'{o}}zefowicz, and
  Bengio]{textvae}
Bowman, Samuel~R., Vilnis, Luke, Vinyals, Oriol, Dai, Andrew~M.,
  J{\'{o}}zefowicz, Rafal, and Bengio, Samy.
\newblock Generating sentences from a continuous space.
\newblock In \emph{Proceedings of the {SIGNLL}'16}, pp.\  10--21, 2016.
\newblock URL \url{https://arxiv.org/abs/1511.06349}.

\bibitem[Burda et~al.(2015)Burda, Grosse, and
  Salakhutdinov]{burda2015importance}
Burda, Yuri, Grosse, Roger, and Salakhutdinov, Ruslan.
\newblock Importance weighted autoencoders.
\newblock \emph{CoRR}, abs/1509.00519, 2015.
\newblock URL \url{http://arxiv.org/abs/1509.00519}.

\bibitem[Cho et~al.(2014)Cho, van Merrienboer, Gulcehre, Bougares, Schwenk, and
  Bengio]{cho2014learning}
Cho, Kyunghyun, van Merrienboer, Bart, Gulcehre, Caglar, Bougares, Fethi,
  Schwenk, Holger, and Bengio, Yoshua.
\newblock Learning phrase representations using {RNN} encoder-decoder for
  statistical machine translation.
\newblock \emph{CoRR}, abs/1406.1078, 2014.
\newblock URL \url{http://arxiv.org/abs/1406.1078}.

\bibitem[Gehring et~al.(2017)Gehring, Auli, Grangier, Yarats, and
  Dauphin]{JonasFaceNet2017}
Gehring, Jonas, Auli, Michael, Grangier, David, Yarats, Denis, and Dauphin,
  Yann~N.
\newblock Convolutional sequence to sequence learning.
\newblock \emph{CoRR}, abs/1705.03122, 2017.
\newblock URL \url{http://arxiv.org/abs/1705.03122}.

\bibitem[Glorot \& Bengio(2010)Glorot and Bengio]{glorot2010understanding}
Glorot, Xavier and Bengio, Yoshua.
\newblock Understanding the difficulty of training deep feedforward neural
  networks.
\newblock In \emph{Proceedings of the Thirteenth International Conference on
  Artificial Intelligence and Statistics}, pp.\  249--256, 2010.

\bibitem[Grathwohl et~al.(2017)Grathwohl, Choi, Wu, Roeder, and
  Duvenaud]{grathwohl2017backpropagation}
Grathwohl, Will, Choi, Dami, Wu, Yuhuai, Roeder, Geoff, and Duvenaud, David.
\newblock Backpropagation through the void: Optimizing control variates for
  black-box gradient estimation.
\newblock \emph{arXiv preprint arXiv:1711.00123}, 2017.

\bibitem[Gu et~al.(2017)Gu, Bradbury, Xiong, Li, and Socher]{nonautoregnmt}
Gu, Jiatao, Bradbury, James, Xiong, Caiming, Li, Victor~O.K., and Socher,
  Richard.
\newblock Non-autoregressive neural machine translation.
\newblock \emph{CoRR}, abs/1711.02281, 2017.
\newblock URL \url{http://arxiv.org/abs/1711.02281}.

\bibitem[Hinton \& Salakhutdinov(2006)Hinton and Salakhutdinov]{reddim}
Hinton, Geoffrey~E. and Salakhutdinov, Ruslan.
\newblock Reducing the dimensionality of data with neural networks.
\newblock \emph{Science}, 313\penalty0 (5786):\penalty0 504--507, 2006.

\bibitem[Hochreiter \& Schmidhuber(1997)Hochreiter and
  Schmidhuber]{hochreiter1997}
Hochreiter, Sepp and Schmidhuber, J{\"u}rgen.
\newblock Long short-term memory.
\newblock \emph{Neural computation}, 9\penalty0 (8):\penalty0 1735--1780, 1997.

\bibitem[Huang \& Jack(1989)Huang and Jack]{huang1989unified}
Huang, XD and Jack, MA.
\newblock Unified techniques for vector quantization and hidden markov modeling
  using semi-continuous models.
\newblock In \emph{Acoustics, Speech, and Signal Processing, 1989. ICASSP-89.,
  1989 International Conference on}, pp.\  639--642. IEEE, 1989.

\bibitem[Huang et~al.(1989)Huang, Hon, and Lee]{huang1989multiple}
Huang, XD, Hon, Hsiao-Wuen, and Lee, Kai-Fu.
\newblock Multiple codebook semi-continuous hidden markov models for
  speaker-independent continuous speech recognition.
\newblock 1989.

\bibitem[Jang et~al.(2016)Jang, Gu, and Poole]{gs1}
Jang, Eric, Gu, Shixiang, and Poole, Ben.
\newblock Categorical reparameterization with gumbel-softmax.
\newblock \emph{CoRR}, abs/1611.01144, 2016.
\newblock URL \url{http://arxiv.org/abs/1611.01144}.

\bibitem[Jegou et~al.(2011)Jegou, Douze, and Schmid]{jegou2011product}
Jegou, Herve, Douze, Matthijs, and Schmid, Cordelia.
\newblock Product quantization for nearest neighbor search.
\newblock \emph{IEEE transactions on pattern analysis and machine
  intelligence}, 33\penalty0 (1):\penalty0 117--128, 2011.

\bibitem[Kaiser \& Bengio(2016)Kaiser and Bengio]{extendedngpu}
Kaiser, {\L}ukasz and Bengio, Samy.
\newblock Can active memory replace attention?
\newblock In \emph{Advances in Neural Information Processing Systems}, pp.\
  3781--3789, 2016.
\newblock URL \url{https://arxiv.org/abs/1610.08613}.

\bibitem[Kaiser \& Bengio(2018)Kaiser and Bengio]{isemhash}
Kaiser, \L{}ukasz and Bengio, Samy.
\newblock Discrete autoencoders for sequence models.
\newblock \emph{CoRR}, abs/1801.09797, 2018.
\newblock URL \url{http://arxiv.org/abs/1801.09797}.

\bibitem[Kaiser \& Sutskever(2016)Kaiser and Sutskever]{neural_gpu}
Kaiser, \L{}ukasz and Sutskever, Ilya.
\newblock Neural {GPU}s learn algorithms.
\newblock In \emph{International Conference on Learning Representations
  ({ICLR})}, 2016.
\newblock URL \url{https://arxiv.org/abs/1511.08228}.

\bibitem[Kaiser et~al.(2017)Kaiser, Gomez, and Chollet]{slicenet}
Kaiser, \L{}ukasz, Gomez, Aidan~N., and Chollet, Francois.
\newblock Depthwise separable convolutions for neural machine translation.
\newblock \emph{CoRR}, abs/1706.03059, 2017.
\newblock URL \url{http://arxiv.org/abs/1706.03059}.

\bibitem[Kalchbrenner \& Blunsom(2013)Kalchbrenner and
  Blunsom]{KalchbrennerB13}
Kalchbrenner, Nal and Blunsom, Phil.
\newblock Recurrent continuous translation models.
\newblock In \emph{Proceedings {EMNLP} 2013}, pp.\  1700--1709, 2013.
\newblock URL \url{http://www.aclweb.org/anthology/D13-1176}.

\bibitem[Kalchbrenner et~al.(2016)Kalchbrenner, Espeholt, Simonyan, van~den
  Oord, Graves, and Kavukcuoglu]{NalBytenet2017}
Kalchbrenner, Nal, Espeholt, Lasse, Simonyan, Karen, van~den Oord, Aaron,
  Graves, Alex, and Kavukcuoglu, Koray.
\newblock Neural machine translation in linear time.
\newblock \emph{CoRR}, abs/1610.10099, 2016.
\newblock URL \url{http://arxiv.org/abs/1610.10099}.

\bibitem[Kingma \& Welling(2013)Kingma and Welling]{vae}
Kingma, Diederik~P. and Welling, Max.
\newblock Auto-encoding variational bayes.
\newblock \emph{CoRR}, abs/1312.6114, 2013.
\newblock URL \url{http://arxiv.org/abs/1312.6114}.

\bibitem[Lam(2018)]{word2bits}
Lam, Maximilian.
\newblock {Word2Bits} -- quantized word vectors.
\newblock \emph{CoRR}, abs/1803.05651, 2018.
\newblock URL \url{http://arxiv.org/abs/1803.05651}.

\bibitem[Lee et~al.(1989)Lee, Hon, Hwang, Mahajan, and Reddy]{lee1989sphinx}
Lee, K-F, Hon, H-W, Hwang, M-Y, Mahajan, Sanjoy, and Reddy, Raj.
\newblock The sphinx speech recognition system.
\newblock In \emph{Acoustics, Speech, and Signal Processing, 1989. ICASSP-89.,
  1989 International Conference on}, pp.\  445--448. IEEE, 1989.

\bibitem[Liu et~al.(2018)Liu, Saleh, Pot, Goodrich, Sepassi, Kaiser, and
  Shazeer]{wikipedia}
Liu, Peter~J., Saleh, Mohammad, Pot, Etienne, Goodrich, Ben, Sepassi, Ryan,
  Kaiser, Lukasz, and Shazeer, Noam.
\newblock Generating wikipedia by summarizing long sequences.
\newblock \emph{CoRR}, abs/1801.10198, 2018.
\newblock URL \url{http://arxiv.org/abs/1801.10198}.

\bibitem[Maddison et~al.(2016)Maddison, Mnih, and Teh]{gs2}
Maddison, Chris~J., Mnih, Andriy, and Teh, Yee~Whye.
\newblock The concrete distribution: {A} continuous relaxation of discrete
  random variables.
\newblock \emph{CoRR}, abs/1611.00712, 2016.
\newblock URL \url{http://arxiv.org/abs/1611.00712}.

\bibitem[Meng et~al.(2015)Meng, Lu, Wang, Li, Jiang, and Liu]{MengLWLJL15}
Meng, Fandong, Lu, Zhengdong, Wang, Mingxuan, Li, Hang, Jiang, Wenbin, and Liu,
  Qun.
\newblock Encoding source language with convolutional neural network for
  machine translation.
\newblock In \emph{ACL}, pp.\  20--30, 2015.
\newblock URL \url{https://arxiv.org/abs/1503.01838}.

\bibitem[Mnih \& Gregor(2014)Mnih and Gregor]{mnih2014neural}
Mnih, Andriy and Gregor, Karol.
\newblock Neural variational inference and learning in belief networks.
\newblock \emph{CoRR}, abs/1402.0030, 2014.
\newblock URL \url{http://arxiv.org/abs/1402.0030}.

\bibitem[Mnih \& Rezende(2016)Mnih and Rezende]{mnih2016variational}
Mnih, Andriy and Rezende, Danilo.
\newblock Variational inference for monte carlo objectives.
\newblock In \emph{International Conference on Machine Learning}, pp.\
  2188--2196, 2016.

\bibitem[Norouzi \& Fleet(2013)Norouzi and Fleet]{norouzi2013cartesian}
Norouzi, Mohammad and Fleet, David~J.
\newblock Cartesian k-means.
\newblock In \emph{Computer Vision and Pattern Recognition (CVPR), 2013 IEEE
  Conference on}, pp.\  3017--3024. IEEE, 2013.

\bibitem[Peinado et~al.(1996)Peinado, Segura, Rubio, Garcia, and
  P{\'e}rez]{peinado1996discriminative}
Peinado, Antonio~M, Segura, Jos{\'e}~C, Rubio, Antonio~J, Garcia, Pedro, and
  P{\'e}rez, Jos{\'e}~L.
\newblock Discriminative codebook design using multiple vector quantization in
  hmm-based speech recognizers.
\newblock \emph{IEEE transactions on speech and audio processing}, 4\penalty0
  (2):\penalty0 89--95, 1996.

\bibitem[Rezende et~al.(2014)Rezende, Mohamed, and
  Wierstra]{rezende2014stochastic}
Rezende, Danilo~Jimenez, Mohamed, Shakir, and Wierstra, Daan.
\newblock Stochastic backpropagation and approximate inference in deep
  generative models.
\newblock \emph{CoRR}, abs/1401.4082, 2014.
\newblock URL \url{http://arxiv.org/abs/1401.4082}.

\bibitem[Rogina \& Waibel(1994)Rogina and Waibel]{rogina1994learning}
Rogina, Ivica and Waibel, Alex.
\newblock Learning state-dependent stream weights for multi-codebook hmm speech
  recognition systems.
\newblock In \emph{Acoustics, Speech, and Signal Processing, 1994. ICASSP-94.,
  1994 IEEE International Conference on}, volume~1, pp.\  I--217. IEEE, 1994.

\bibitem[Salakhutdinov \& Hinton(2009{\natexlab{a}})Salakhutdinov and
  Hinton]{deepbm}
Salakhutdinov, Ruslan and Hinton, Geoffrey~E.
\newblock Deep {B}oltzmann machines.
\newblock In \emph{Proceedings of {AISTATS}'09}, pp.\  448--455,
  2009{\natexlab{a}}.

\bibitem[Salakhutdinov \& Hinton(2009{\natexlab{b}})Salakhutdinov and
  Hinton]{semhash}
Salakhutdinov, Ruslan and Hinton, Geoffrey~E.
\newblock Semantic hashing.
\newblock \emph{Int. J. Approx. Reasoning}, 50\penalty0 (7):\penalty0 969--978,
  2009{\natexlab{b}}.

\bibitem[Shu \& Nakayama(2018)Shu and Nakayama]{shu2018compressing}
Shu, Raphael and Nakayama, Hideki.
\newblock Compressing word embeddings via deep compositional code learning.
\newblock In \emph{International Conference on Learning Representations}, 2018.
\newblock URL \url{https://openreview.net/forum?id=BJRZzFlRb}.

\bibitem[Subakan et~al.(2018)Subakan, Koyejo, and Smaragdis]{aepredict}
Subakan, Cem, Koyejo, Oluwasanmi, and Smaragdis, Paris.
\newblock Learning the base distribution in implicit generative models.
\newblock \emph{CoRR}, abs/1803.04357, 2018.
\newblock URL \url{http://arxiv.org/abs/1803.04357}.

\bibitem[Sutskever et~al.(2014)Sutskever, Vinyals, and Le]{sutskever14}
Sutskever, Ilya, Vinyals, Oriol, and Le, Quoc~V.
\newblock Sequence to sequence learning with neural networks.
\newblock In \emph{Advances in Neural Information Processing Systems}, pp.\
  3104--3112, 2014.
\newblock URL \url{http://arxiv.org/abs/1409.3215}.

\bibitem[Tucker et~al.(2017)Tucker, Mnih, Maddison, Lawson, and
  Sohl-Dickstein]{tucker2017rebar}
Tucker, George, Mnih, Andriy, Maddison, Chris~J, Lawson, John, and
  Sohl-Dickstein, Jascha.
\newblock Rebar: Low-variance, unbiased gradient estimates for discrete latent
  variable models.
\newblock In \emph{Advances in Neural Information Processing Systems}, pp.\
  2624--2633, 2017.

\bibitem[van~den Oord et~al.(2017)van~den Oord, Vinyals, and
  Kavukcuoglu]{vqvae}
van~den Oord, A{\"{a}}ron, Vinyals, Oriol, and Kavukcuoglu, Koray.
\newblock Neural discrete representation learning.
\newblock \emph{CoRR}, abs/1711.00937, 2017.
\newblock URL \url{http://arxiv.org/abs/1711.00937}.

\bibitem[van~den Oord et~al.(2016)van~den Oord, Dieleman, Zen, Simonyan,
  Vinyals, Graves, Kalchbrenner, Senior, and Kavukcuoglu]{wavenet}
van~den Oord, Aäron, Dieleman, Sander, Zen, Heiga, Simonyan, Karen, Vinyals,
  Oriol, Graves, Alexander, Kalchbrenner, Nal, Senior, Andrew, and Kavukcuoglu,
  Koray.
\newblock {WaveNet}: A generative model for raw audio.
\newblock \emph{CoRR}, abs/1609.03499, 2016.
\newblock URL \url{http://arxiv.org/abs/1609.03499}.

\bibitem[Vaswani et~al.(2017)Vaswani, Shazeer, Parmar, Uszkoreit, Jones, Gomez,
  Kaiser, and Polosukhin]{transformer}
Vaswani, Ashish, Shazeer, Noam, Parmar, Niki, Uszkoreit, Jakob, Jones, Llion,
  Gomez, Aidan~N., Kaiser, Lukasz, and Polosukhin, Illia.
\newblock Attention is all you need.
\newblock \emph{CoRR}, 2017.
\newblock URL \url{http://arxiv.org/abs/1706.03762}.

\bibitem[Vaswani et~al.(2018)Vaswani, Parmar, Uszkoreit, Shazeer, and
  Kaiser]{imagetrans}
Vaswani, Ashish, Parmar, Niki, Uszkoreit, Jakob, Shazeer, Noam, and Kaiser,
  Lukasz.
\newblock Image transformer, 2018.
\newblock URL \url{https://openreview.net/forum?id=r16Vyf-0-}.

\bibitem[Vincent et~al.(2010)Vincent, Larochelle, Lajoie, Bengio, and
  Manzagol]{stackeddenoising}
Vincent, Pascal, Larochelle, Hugo, Lajoie, Isabelle, Bengio, Yoshua, and
  Manzagol, Pierre{-}Antoine.
\newblock Stacked denoising autoencoders: Learning useful representations in a
  deep network with a local denoising criterion.
\newblock \emph{Journal of Machine Learning Research}, 11:\penalty0 3371--3408,
  2010.

\bibitem[Vinyals et~al.(2015)Vinyals, Kaiser, Koo, Petrov, Sutskever, and
  Hinton]{KVparse15}
Vinyals, Kaiser, Koo, Petrov, Sutskever, and Hinton.
\newblock Grammar as a foreign language.
\newblock In \emph{Advances in Neural Information Processing Systems}, 2015.
\newblock URL \url{http://arxiv.org/abs/1412.7449}.

\bibitem[Williams(1992)]{williams1992simple}
Williams, Ronald~J.
\newblock Simple statistical gradient-following algorithms for connectionist
  reinforcement learning.
\newblock In \emph{Reinforcement Learning}, pp.\  5--32. Springer, 1992.

\bibitem[Yang et~al.(2017)Yang, Hu, Salakhutdinov, and
  Berg{-}Kirkpatrick]{textvae2}
Yang, Zichao, Hu, Zhiting, Salakhutdinov, Ruslan, and Berg{-}Kirkpatrick,
  Taylor.
\newblock Improved variational autoencoders for text modeling using dilated
  convolutions.
\newblock In \emph{Proceedings of {ICML}'17}, pp.\  3881--3890, 2017.

\end{thebibliography}
\bibliographystyle{icml2018}

\end{document}